\documentclass[letterpaper, 10 pt, conference]{ieeeconf}  

\IEEEoverridecommandlockouts                              

\usepackage[font=small]{caption}
\usepackage{graphicx} 
\usepackage{tabularx}
\usepackage{times} 
\usepackage{amsmath} 
\usepackage{amssymb}  
\usepackage{comment}
\usepackage{url}
\usepackage{subfigure}
\usepackage{multirow}
\usepackage{colortbl}
\usepackage{booktabs}
\usepackage[table,xcdraw]{xcolor}
\usepackage[normalem]{ulem}
\newcolumntype{x}[1]{>{\centering\arraybackslash}p{#1pt}}
\newcolumntype{y}[1]{>{\raggedright\arraybackslash}p{#1pt}}
\newcolumntype{z}[1]{>{\raggedleft\arraybackslash}p{#1pt}}
\usepackage{xcolor}

\makeatletter
\let\NAT@parse\undefined
\makeatother
\usepackage[pagebackref=false,breaklinks=true,letterpaper=true,colorlinks,bookmarks=false]{hyperref}
\newcommand{\tablestyle}[2]{\setlength{\tabcolsep}{#1}\renewcommand{\arraystretch}{#2}\centering\footnotesize}
\useunder{\uline}{\ul}{}

\begin{document}
\title{\LARGE \bf
DHP-Mapping: A Dense Panoptic Mapping System with  Hierarchical World Representation and Label Optimization Techniques
}

\author{Tianshuai Hu$^{1}$, Jianhao Jiao$^{3}$, Yucheng Xu$^{4}$, Hongji Liu$^{1}$, Sheng Wang$^{1}$, and Ming Liu$^{2}$
\thanks{
  $^{1}$The Hong Kong University of Science and Technology, China.
  {\tt\small \{thuaj,hliucq,swangei\}}@connect.ust.hk.\ 
  $^{2}$The Hong Kong University of Science and Tehcnology (Guangzhou), China.
  {\tt\small eelium}@hkust-gz.edu.cn\ 
  $^{3}$University College London, UK
  {\tt\small ucacjji}@ucl.ac.uk\ 
  $^{4}$University of Edinburgh, UK.
  {\tt\small yucheng.xu}@ed.ac.uk.
  }
}

\maketitle
\thispagestyle{empty}
\pagestyle{empty}

\begin{abstract}
Maps provide robots with crucial environmental knowledge, thereby enabling them to perform interactive tasks effectively. 
Easily accessing accurate abstract-to-detailed geometric and semantic concepts from maps is crucial for robots to make informed and efficient decisions.
To comprehensively model the environment and effectively manage the map data structure, we propose DHP-Mapping, a dense mapping system that utilizes multiple Truncated Signed Distance Field (TSDF) submaps and panoptic labels to hierarchically model the environment. The output map is able to maintain both voxel- and submap-level metric and semantic information.
Two modules are presented to enhance the mapping efficiency and label consistency:
(\text{1}) an inter-submaps label fusion strategy to eliminate duplicate points across submaps and
(\text{2}) a conditional random field (CRF) based approach to enhance panoptic labels through object label comprehension and contextual information.
We conducted experiments with two public datasets including indoor and outdoor scenarios.
Our system performs comparably to state-of-the-art (SOTA) methods across geometry and label accuracy evaluation metrics. The experiment results highlight the effectiveness and scalability of our system, as it is capable of constructing precise geometry and maintaining consistent panoptic labels.
Our code is publicly available at ~\url{https://github.com/hutslib/DHP-Mapping}.
\end{abstract}
\section{Introduction}
\label{sec:intro}
\subsection{Motivation}
In modern autonomous robot systems, maps serve as representations of semantic and geometry information of the environment, which are essential for robots to interact with the real world and operate based on surrounding semantic information.
In this paper, we identify two key features that are important for an ideal dense 3D semantic map, which, despite their significance, have not been fully realized in existing mapping systems:

(1)~\textbf{Comprehensive and accurate Labels}. A completed dense semantic 3D map should capture precise foreground and background information at both semantic and instance levels. This ensures the delivery of reliable and unique information for specific tasks. 
Considering different types of objects, foreground entities involve active objects that robots might interact with~\cite
{rana2023sayplan,gu2023conceptgraphs}, while background entities provide meaningful static constraints for robot localization~\cite{schonberger2018semantic,millane2021freetures} and navigation~\cite{maturana2018real}. 
Regarding levels of information, semantic information categorizes objects and outlines their properties, equipping robots with the knowledge to determine appropriate actions for different objects, whereas instance information helps robots distinguish the exact object to interact with.

(2)~\textbf{Hierarchical Data Structure}. 
The data structure should be able to efficiently manage and store information that spans multiple levels of detail and complexity, thus enabling robots to quickly retrieve necessary information during task execution. 
Stacking environmental geometric and semantic data in a flat structure is an intuitive way to maintain a metric-semantic map. However, it might suffer from several limitations such as inefficient data storage, complex map modification processes, and an inability to capture the abstract attributes that are essential for certain robot tasks. For example, global planning~\cite{rana2023sayplan,galindo2008robot} often requires instance-level data and relationships among objects, while local planning~\cite{zhao2019semantic,han2019fiesta} and collision detection~\cite{huang2023voxposer} prefer dense local information and object property. 
Hierarchical scene representation is a promising approach to address these limitations, as it offers a representation that facilitates easy data access and encompasses multi-level information~(e.g. instance-level, voxel-level). 

\begin{figure}[t]
    \centering
    \includegraphics[width=0.85\linewidth]{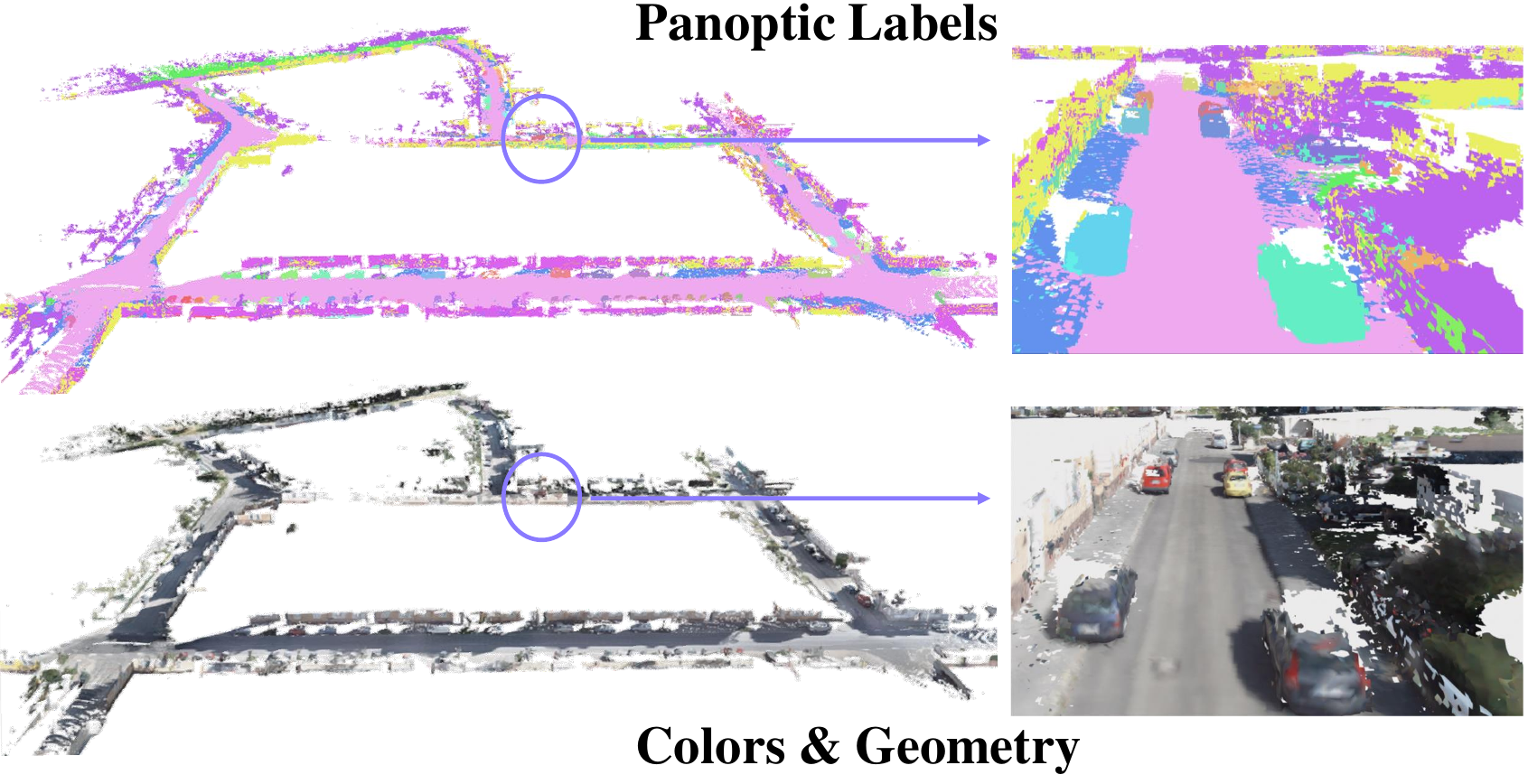}
    \caption{DHP-Mapping portrays the scenario through a collection of TSDF submaps, each submap represents either a foreground thing or background stuff together with semantic class and instance ID.}
    \label{fig:main_pic}
    \vspace{-1em}
\end{figure}

\begin{figure*}[t!]
    \centering
    \includegraphics[width=0.95\linewidth]{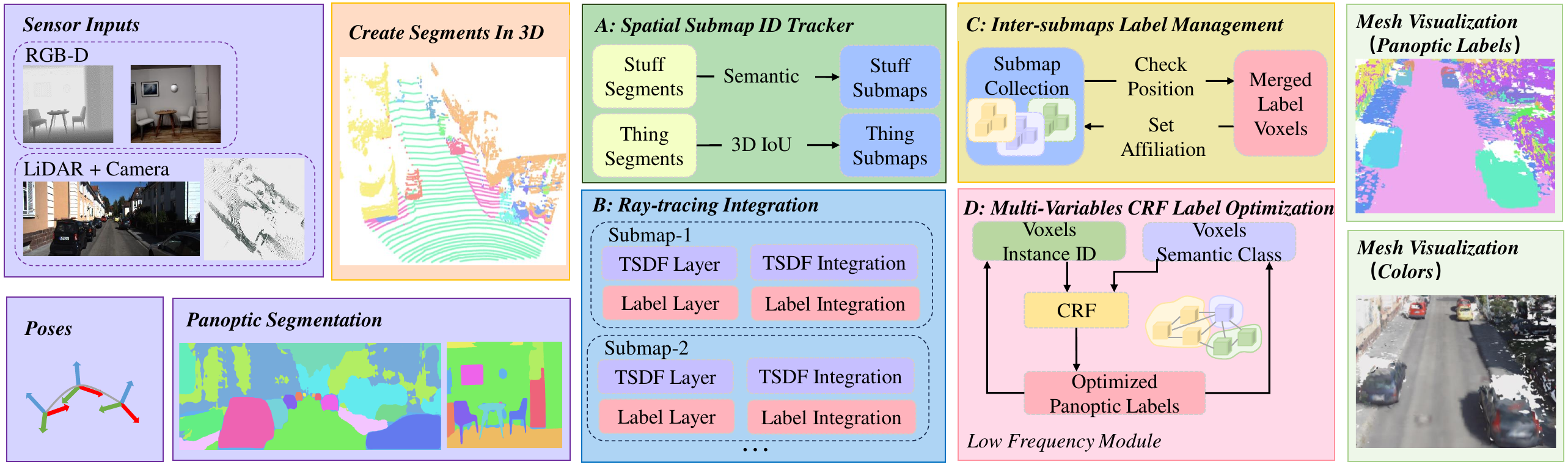}
    \caption{
Our system converts sensor data, poses, and panoptic segmentation into point segments.
(A): A data association process utilizes semantic categories and 3D IoU metrics to assign a submap ID to each segment. 
(B): Segment information is integrated into the assigned submap's TSDF and label layers through ray-tracing. 
(C): To prevent submap overlap, voxels occupying the same space are identified and their information is fused, ensuring each submap exclusively represents an object.
(D): A CRF algorithm enhances the precision of semantic and instance labels by encouraging label consistency among voxels exhibiting similar color and geometric features.
}
\label{fig:pipeline}
\vspace{-1.5em}
\end{figure*}

\subsection{Related Works}
Traditional segmentation-based dense mapping systems can be classified into two different categories: dense semantic labeling \cite{Rosinol20icra-Kimera,yang2017semantic,mccormac2017semanticfusion,Jiao2023OnlineMetricSemantic} and instance-orientated \cite{mccormac2018fusion++,mascaro2022volumetric,salas2013slam++,wang2021dsp} approaches.
Typically, semantic mapping methods~\cite{Rosinol20icra-Kimera,mccormac2017semanticfusion,yang2017semantic,Jiao2023OnlineMetricSemantic} can reconstruct scenes with dense semantic labels but lack the instance information. The instance mapping methods~\cite{mccormac2018fusion++, mascaro2022volumetric, salas2013slam++,wang2021dsp} only reconstruct the foreground while ignoring the background. However, none of them manages to integrate information at both the semantic and instance levels while incorporating both foreground and background elements, which leads to a failure to create a comprehensive map.

A possible solution for achieving more comprehensive 3D scene interpretation is to take advantage of the progress made in panoptic segmentation ~\cite{kirillov2019panoptic}. 
Recent works extend panoptic segmentation to scanned point clouds~\cite{jayasumana2019bipartite,pham2019jsis3d}, and videos~\cite{kim2020video} aiming to considerate the temporal dimension and improve overall scene understanding. However, most of these methods are offline, requiring pre-collected data. 
Other works~\cite{narita2019panopticfusion,yang2021tupper} extend the concept of panoptic segmentation to online mapping.
However, these maps are built in a flat manner, where dense 3D maps add information to each storage unit without taking into account any connections or relationships among them. 
The lack of hierarchical world representation causes challenges in maintaining the constructed maps and executing high-level tasks efficiently.

In the field of hierarchical 3D scene modeling, studies such as~\cite{hughes2022hydra, hughes2023foundations} aim to improve indoor scene representations beyond metric-semantic map by integrating high-level attributes such as objects, rooms, and free spaces into a scene graph. 
Here, nodes represent spatial and semantic concepts, with edges depicting the relationships among them, forming a comprehensive and organized representation of the map.
Another particular method in~\cite{schmid2022panoptic} adopts an object-level submap-based representation, dividing the environment into smaller, manageable sections that can be processed, updated, and removed independently. The final label of each voxel is predicted by merging the panoptic segmentation results of multiple frames. 
Taking advantage of the hierarchical structure, these aforementioned methods effectively reduce memory consumption and accelerate the information retrieval process. However, these approaches have limitations: they neither leverage the full scope of scene information for enhancing label accuracy nor employ hierarchical frameworks to refine label estimations for each object which may cause inconsistent labels across mapping.

\subsection{Contributions}
To address the limitations of existing mapping methods and meet nowadays requirements of robotic applications, we introduce DHP-Mapping (\textbf{Dense, Hierarchical, and Panoptic}), a novel 3D mapping system. This system distinguishes itself by constructing and managing a hierarchical structure of multiple TSDF submaps with optimized consistent panoptic labels. 
Our approach is open-source at \color{blue}\href{https://github.com/hutslib/DHP-Mapping}{https://github.com/hutslib/DHP-Mapping}\color{black}.
We summarize our main contributions as follows:
\begin{itemize}
    \item 
    We build a 3D mapping system, which hierarchically represents 3D scenes with multiple TSDF submaps and panoptic labels. Our system features an inter-submap label management module designed to fuse labels across submaps and eliminate overlaps between submaps. This design guarantees the integrity of the hierarchical structure and boosts data retrieval efficiency.
    
    \item 
    We introduce a conditional random field (CRF) based optimization module. This strategy leverages both geometric and color information throughout the scene to enhance the consistency and accuracy of labels.
    
    \item 
    We conducted both indoor and outdoor experiments with different sensor settings, and compared our system with SOTA algorithms. The results show that our method performs comparable to SOTA algorithms in terms of label accuracy and geometry precision and exhibits more scalability.
\end{itemize}
\section{Methodology} \label{Sec:pipeline}
The pipeline of our system is illustrated in Fig. \ref{fig:pipeline}.
Given the sensor inputs, robot poses and panoptic segmentation results, the primary objective of our mapping system is to construct a hierarchical map data structure and also incorporate comprehensive semantic information. Achieving the goal is twofold: (\textit{i}) Reconstruct the environment through a collection of TSDF submaps, denoted as $\mathcal{G}=\{\mathcal{M}_{1}, \dots, \mathcal{M}_N\}$. (\textit{ii}) For each submap $\mathcal{M}_i$, our system estimates a pair of random variables $^{i}L~=~\left( ^{i}X_S,\; ^{i}X_I\right)$ as its panoptic label, where $^{i}X_S$ and $^{i}X_I$ stand for the semantic class label and the instance ID label of $\mathcal{M}_i$ respectively. 

In this section, we first introduce the hierarchical data structure of our mapping system.
After that, we present the core modules in our mapping systems individually.

\subsection{Multiple TSDF Submaps with Panoptic Labels Data Structure} \label{sec:data structure}
\begin{figure}[t]
    \centering
    \includegraphics[width=0.95\linewidth]{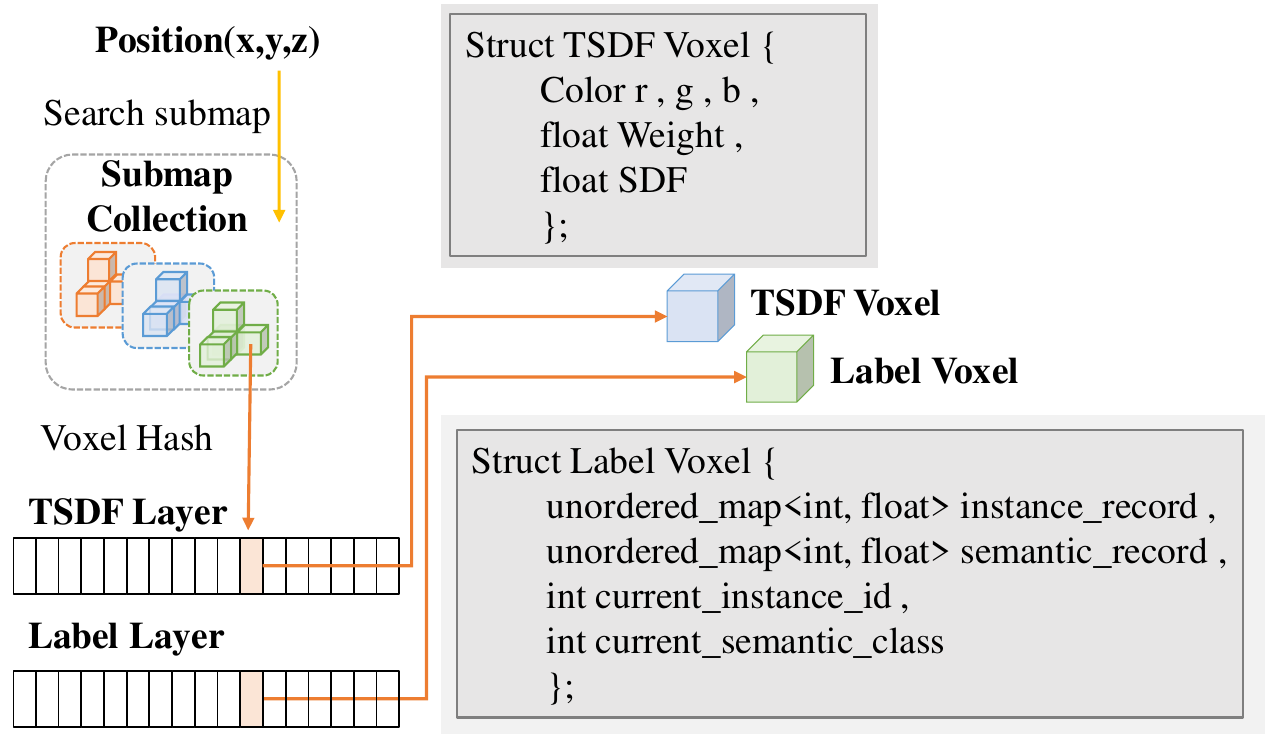}
    \caption{To query a point in the world space, the first step is to identify its corresponding submap via submap bounding volumes. Then a hash function can be employed to query the TSDF and label voxels within each layer separately. }
    \label{fig:data_structure}
    \vspace{-1.5em}
\end{figure}

As shown in Fig. \ref{fig:data_structure}, the environment is represented as a collection of submaps. Each submap consists of a TSDF layer and a label layer to separate the storage of geometry and label information. This is achieved by extending Voxblox \cite{oleynikova2017voxblox} with an extra layer for label storage. Each voxel within the TSDF layer stores its own TSDF value, weight, and color. Similarly, each voxel in the label layer is associated with a specific semantic class and instance ID.
During the mapping process, panoptic information for a voxel from previous frames is collected and their counts are recorded.

Apart from the voxel-level value storage, the instance-level information is maintained via the submap collection. Each submap has an indexing ID and a panoptic label. The submap-level geometric information is expressed as a sphere-shaped bounding volume that encompasses all voxels.

Notable, to establish and maintain the hierarchical connection between submaps and voxels, each submap is utilized to represent a unique instance and group voxels inside it as a unit. Thus, upon completion of the mapping process, each voxel is stored only in the submap that shares the same instance ID as it. This strategy allows for the omission of individual voxel label information,  markedly reducing storage consumption. Besides, this hierarchical storage strategy can improve the efficiency of data retrieval and modification. To acquire voxels associated with a specific object, the hierarchical structure enables direct retrieval via indexing to the corresponding submap, bypassing the need for exhaustive traversal and label verification of each voxel.

\subsection{Spatial Submap ID tracker}\label{sec:spatialidtracker}
The input data, containing the poses, panoptic segmentation, color, and depth information, is initially processed and grouped based on the frame-wise instance ID, resulting in a collection of segments.
However, directly integrating each segment into a submap collection is problematic, as frame-wise segmentation does not guarantee the consistency of instance IDs across frames. Thus, our spatial submap ID tracker aims to find a relation to associate each segment with a submap already in the submap collection.

Based on the panoptic label of a segment, different strategies are implemented to associate the segment with a submap in the collection. A segment with a background category gets a submap ID if the same semantic class submap exists. A segment with a foreground category is tracked by ray-tracing its points into the world coordinate and selecting the submap with the highest IoU among all foreground-class submaps in the collection. For unmatched segments with enough points, a new submap will be created.

Unlike previous methods~\cite{schmid2022panoptic,mccormac2018fusion++,narita2019panopticfusion}, which project the map iso-surface points back to the image plane, our method directly queries the voxel position in submap collection and calculates 3D tracking statistics, which can leverage the rich 3D information, incorporating an extra depth dimension to identify entities that represent the same object. Using the panoptic labels as a prior, different strategies are employed for stuff and thing class segments to prevent merging foreground class segments into background submaps and reduce the spurious creation of same-class background submaps.

\begin{figure}[t]
    \centering
    \includegraphics[width=1.0\linewidth]{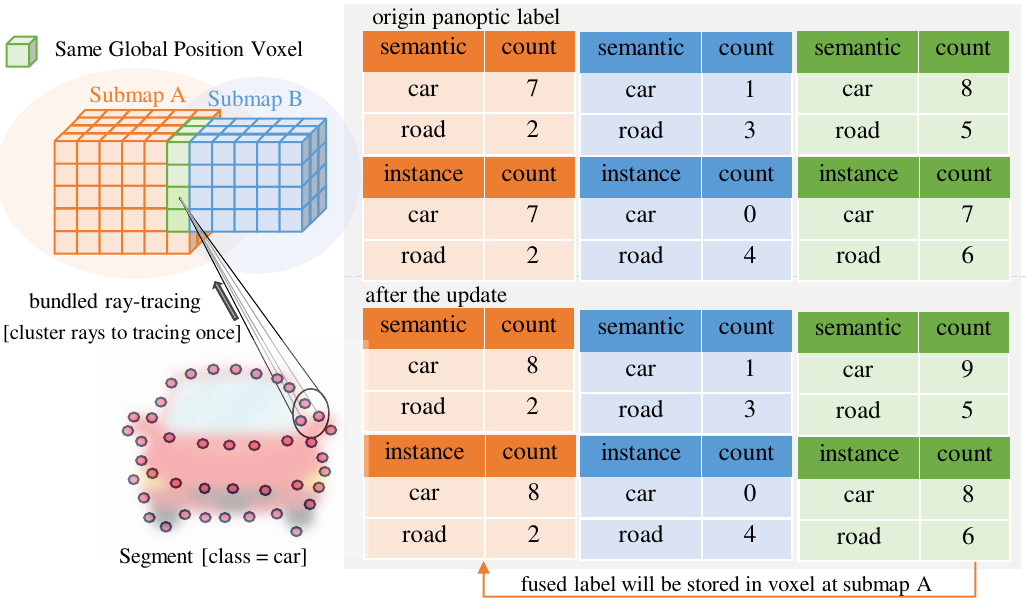}
     \vspace{-1.5em}
    \caption{The left part of this figure shows bundled ray-tracing of points in a segment. The segment is classified with semantic class \textbf{car} and is associated with submap \textbf{A}. It also illustrates that a spatial position can be covered by more than one submap as indicated by \textcolor[RGB]{126,180,66}{voxels marked green}. The \textcolor[RGB]{237,125,49}{orange tables} details the update process for a voxel in submap A by the bundled ray. Specifically, the semantic count of \textbf{car} and instance ID count of \textbf{A} are both increased by one. The \textcolor[RGB]{112,168,218}{blue tables} present the semantic and instance records for a voxel within submap B, which occupies the same spatial position as the voxel in submap A. The \textcolor[RGB]{126,180,66}{green tables} display the label information for this spatial position, showcasing merged data results from submap A and B. It's important to note that the merged label will be solely stored in a voxel of submap A after the process described in \ref{sec:mapmanagement}.}
    \label{fig:update}
    \vspace{-1.5em}
\end{figure}

\subsection{Ray-tracing Integration}\label{sec:raycasteringintegration}
Based on the tracking ID, segment information updates the corresponding submap. As illustrated in \ref{sec:data structure}, geometry and label information are stored separately and need to be integrated into corresponding layers. The integration strategy for TSDF layers is similar to the original Voxblox~\cite{oleynikova2017voxblox}.
Owing to changes in observation angles and imperfections in the segmentation network's output, panoptic label predictions for the same location can vary across frames. Thus, the label integration process aims to get together all label information from past frames and select the label with the highest probability for each voxel and submap.

As illustrate by the Fig.~\ref{fig:update}, we employ bundled ray-tracing to propagate label information from a segment along the ray to voxels within the truncated distance and contained in its associated submap. Each touched voxel experiences an increment in its count for the segment's semantic category by one. Similarly, the count for the submap instance ID will be incremented by one, indicating that this spatial location is associated with this submap once. The panoptic label for each voxel is obtained by finding the semantic and instance ID that have the maximum count. The label probability is computed as the ratio of the occurrence frequency for a particular semantic/instance label to the total occurrence frequency across all semantic/instance labels. The panoptic label of a submap is determined by calculating the labels with the highest occurrence among all the voxels inside it.

\subsection{Inter-Submaps Label Management}\label{sec:mapmanagement}
Creating submaps from segments can lead to multiple submaps that describe the identical position in 3D space, as depicted in Fig.~\ref{fig:update}. Retriving geometry and label information for a specific position requires searching through all submaps to identify those containing the queried position and then merging their stored values. This process's computational complexity grows linearly with the number of submaps. As the submap collection expands, data retrieving becomes a non-trivial task. Therefore, we propose an inter-submaps label management strategy to ensure that each 3D position is exclusively stored within a single submap.

Submaps not associated with any segments for several consecutive frames are deemed inactive, indicating they are currently out of the field of view. Label management operates on these inactive submaps. It starts by identifying overlapping submaps pairs, denoted as $(\mathcal{M}_{a}, \mathcal{M}_{b})$, through checking intersections among submap bounding volumes. Then, blocks $ \mathcal{B}_{a}$ in $\mathcal{M}_{a}$ that fall within the bounding volume of $\mathcal{M}_{b}$ are retrieved for further detecting voxels occupying identical positions. Upon detecting a voxel $v_{a}$ in $\mathcal{B}_{a}$ that coincides positionally with a voxel $v_{b}$ in $\mathcal{M}_{b}$, we merge the TSDF and label data from $v_{a}$ into $v_{b}$. The merging procedure involves a weighted average of the SDF values and an aggregation of label counts. The panoptic label for voxel $v_{b}$ is then recalculated based on the aggregated label counts, and any data pertaining to $v_{a}$ is deleted. The final step involves reassigning voxels based on instance ID agreement, relocating voxels with mismatched IDs to the correct submap and purging them from current storage.

Notably, unlike previous methods, each unique 3D point will be captured in only one submap. Therefore, given a query point, the corresponding information can be directly and efficiently retrieved from the submap that contains it, as illustrated in Fig.~\ref{fig:data_structure}. This approach helps us better maintaing the hierarchical map structure and ensures the coherence and exclusivity of spatial information in the submaps. 

\subsection{Multi-Variables CRF Label Optimization}\label{sec:crfrefinement}
Directly incorporating segmentation results into the map will only captures the temporal label information from different observations. However, objects exist as cohesive units in reality, necessitating the incorporation of spatial and object-level information to enhance the accuracy of panoptic label estimation. To achieve this, we employ a multi-variable CRF to optimize semantic and instance label. 

The multi-variables CRF executes on inactive submaps and is characterized by a Gibbs energy shown as follows:
\begin{equation}
\begin{aligned}
 P(X_{S},X_{I}|D)=\frac{1}{Z(D)} 
 \exp[-E(X_{S},X_{I}|D)],
\end{aligned}
\end{equation}
where $P(X_{S}, X_{I} |D)$ is the posterior probability of the joint configuration for semantic labels $X_{S}$ and instance labels $X_{I}$, given the information from the voxel data $D$. 
$Z(D)$ is the normalization factor, $E(X_{S}, X_{I}|D)$ is the energy term with variables configuration $(X_{S}, X_{I})$.
The objective of the multi-variables CRF is to find an optimized configuration  $\left(X_{S}, X_{I}\right)$, such that,
\begin{equation}\label{eq:Gibbs_energy}
\begin{aligned}
(X_{S}, X_{I})^* = 
\underset{(X_{S}, X_{I})}{\arg\max}\ E(X_{S}, X_{I}|D),
\end{aligned}
\end{equation}
which is solved using mean field inference algorithm \cite{krahenbuhl2011efficient}.
Hereafter, we omit the conditioning on $D$ in our notation for simplicity.
In this task, we define our energy terms as:
\begin{equation}\label{eq:energy term}
\begin{aligned}
E(X_{S}, X_{I}) = 
\sum_{y}\varphi^{U} \left (^{y}X_S \right ) +  
\sum_{y}&\psi^{U}  \left (^{y}X_I\right )  \\ +   
\sum_{\left(y,z\right),y<z}\varphi^{P} \left (^{y}X_S ,  ^{z}X_S \right ) +
\sum_{\left(y,z\right),y<z}&\psi ^{P} \left( ^{y}X_I , ^{z}X_I \right), 
\end{aligned}
\end{equation}
where the unary term for semantic label $\varphi^{U} \left (^{y}X_S \right )$ and instance label $\psi^{U}  \left (^{y}X_I\right )$ represent the prior cost to assign a label to a voxel $y$, and here we use the negative logarithm of the label probability obtained in Section \ref{sec:raycasteringintegration}. 
 The pairwise terms penalize voxel pairs that have similar colors and proximate center positions, yet with different semantic or instance labels. Inspired by \cite{yang2017semantic}, both of the semantic and instance pairwise terms using a combination of two Gaussian kernels and defined based on the color and position information of the voxels:
\begin{equation}
\begin{aligned}
\varphi^{P} \left( ^{y}X_S , ^{z}X_S \right)  &= \mu\left(^{y}X_S , ^{z}X_S\right)\sum_{m=1,2}\omega_{m}K_{m}(y,z), \\
\psi ^{P} \left( ^{y}X_I , ^{z}X_I \right)  &= \mu\left(^{y}X_I , ^{z}X_I\right)\sum_{m=1,2}\omega_{m}K_{m}(y,z),
\end{aligned}
\end{equation}
where $\mu\left(^{y}X_S , ^{z}X_S\right)$ is a binary label compatibility function. $\mu\left(^{y}X_S , ^{z}X_S\right) = 0$ when $^{y}X_S = {^{z}X_S}$, otherwise $\mu\left(^{y}X_S , ^{z}X_S\right) = 1$. The same applies to  $\mu\left(^{y}X_I , ^{z}X_I\right)$.
The Gaussian kernels are designed to be the same for both the semantic and instance term as follows:
\begin{equation}
    \begin{aligned}
K\left(y,z\right) 
&= \omega_{1} K_{1} \left( y,z\right)+\omega_{2}K_{2} \left( y,z\right)\\
&=  \omega_{1} \exp\left( -\frac{\left | \mathbf{p}_{y}-\mathbf{p}_{z} \right | }{2\theta^{2}_{1} }  -\frac{\left | \mathbf{o}_{y}-\mathbf{o}_{z} \right | }{2\theta^{2}_{2} }  \right) \\
&+ \omega_{2}\exp\left( -\frac{\left | \mathbf{p}_{y}-\mathbf{p}_{z} \right | }{2\theta^{2}_{1} } \right),
    \end{aligned}
\end{equation}
where $\mathbf{p}$ is the center position of a voxel and $\mathbf{o}$ is the color value stored in this voxel.
\section{Experiments and Results}
\label{sec:experiment_and_results}
\begin{table*}[t!]
\centering
\resizebox{0.95\textwidth}{!}{
\tablestyle{4pt}{1.1}
\begin{tabular}{x{35}y{80}x{30}x{30}x{30}x{40}x{40} x{42}x{42}x{42}}
     \toprule
\multirow{2}{*}{\textbf{Dataset}} & \multirow{2}{*}{\textbf{Method}} & \multicolumn{3}{c}{\textbf{Panoptic Metric}}& \multicolumn{2}{c}{\textbf{Semantic Metric}}& \multicolumn{3}{c}{\textbf{Instance Metric}} \\
\cmidrule(lr){3-5} \cmidrule(lr){6-7} \cmidrule(lr){8-10}
& & PQ $\uparrow$ & RQ $\uparrow$ & SQ $\uparrow$& mIoU $\uparrow$& Accuracy $\uparrow$& mAP-(0.3) $\uparrow$ & mAP-(0.4) $\uparrow$ & mAP-(0.5) $\uparrow$ \\
\midrule
 & Panmap & {\ul 0.534} & \textbf{0.804} & 0.246 & \textbf{0.271} & {\ul0.679} & 0.546 & 0.541  & 0.532 \\
\cmidrule(lr){2-2}
 & DHP w/o refine & 0.533 & 0.678 & {\ul0.297} & 0.240 & 0.678 & {\ul0.765}  & {\ul0.761} & {\ul0.752} \\
  \multirow{-3}{*}{\begin{tabular}[c]{@{}c@{}}Semantic\\KITTI\end{tabular}}  & 
  \cellcolor[HTML]{EFEFEF}  DHP & \cellcolor[HTML]{EFEFEF} \textbf{0.568} &  \cellcolor[HTML]{EFEFEF} {\ul0.701} &  \cellcolor[HTML]{EFEFEF} \textbf{0.340} &  \cellcolor[HTML]{EFEFEF} {\ul0.263} & \cellcolor[HTML]{EFEFEF}  \textbf{0.713} & \cellcolor[HTML]{EFEFEF} \textbf{0.769} & \cellcolor[HTML]{EFEFEF} \textbf{0.765}  & \cellcolor[HTML]{EFEFEF} \textbf{0.758}\\
\midrule
 & Panmap & 0.436 & 0.506 & 0.274 & 0.378 & {\ul 0.659} & 0.512 & 0.512  & 0.464\\
\cmidrule(lr){2-2}
 & DHP w/o refine & {\ul0.660} & {\ul0.660} & {\ul0.357} & {\ul0.382} & 0.631 & \textbf{0.690} & { \ul0.667}  & {\ul0.667}\\
  \multirow{-3}{*}{flat}  & 
  \cellcolor[HTML]{EFEFEF}  DHP & \cellcolor[HTML]{EFEFEF} \textbf{0.708}&  \cellcolor[HTML]{EFEFEF} \textbf{0.787} &  \cellcolor[HTML]{EFEFEF} \textbf{0.702} &  \cellcolor[HTML]{EFEFEF} \textbf{0.629} & \cellcolor[HTML]{EFEFEF}  \textbf{0.854} & \cellcolor[HTML]{EFEFEF} \textbf{0.690} & \cellcolor[HTML]{EFEFEF} \textbf{0.690}  & \cellcolor[HTML]{EFEFEF} \textbf{0.690} \\
 \bottomrule
\end{tabular}
}
\caption{Label accuracy comparisons with a SOTA panoptic mapping method and impact of our proposed label refinement modules. This table displays the quantitative evaluation results of our method versus the SOTA panoptic mapping approach, across two datasets. We compare their performances using the Panoptic, Semantic, and Instance Metrics. Additionally, we compare the effect of with/without our label refinement module (Algorithms described in Sec~\ref{sec:mapmanagement} and Sec~\ref{sec:crfrefinement}). We \textbf{bold} the best results and {\ul underline} the second best results. Note: $\textrm{mAP-}(x)$ indicates the evaluation is conducted with $\textrm{IoU} = x$.
}
\label{tab:main-eval}
\vspace{-2em}
\end{table*}

\subsection{Experimental Setup}
\textbf{Dataset}.
We conduct experiments on indoor simulation and outdoor real-world datasets. These datasets have different sensor settings~(RGB-D only and LiDAR-camera), and different levels of scene complexity.

(1)~\textit{flat dataset}~\cite{schmid2022panoptic} is an indoor simulation dataset, which provides RGB-D images, pixel-level ground truth panoptic segmentation labels, and ground truth poses.

(2) \textit{SemanticKITTI dataset}~\cite{behley2019iccv} is an outdoor real-world dataset, which provides RGB images, LiDAR point clouds with ground truth panoptic labels, and ground truth poses. 

\textbf{Ground truth labeled meshes}.
We use dataset-provided ground truth panoptic labels with other sensor inputs, to generate labeled 3D points with consistent panoptic information across frames. These points are directly integrated into a map, creating TSDF maps with voxel-wise panoptic labels. The ground truth labeled meshes are extracted from the TSDF maps using the marching cubes algorithm \cite{lorensen1998marching}. 

\textbf{Metrics}. To evaluate map label accuracy, we use 3D Panoptic Quality (PQ), Segmentation Quality (SQ), and Recognition Quality (RQ) for panoptic labels; 3D mean Intersection over Union (mIoU) and Accuracy for semantic labels; and 3D mean Average Precision (mAP) for instance labels. For map geometry accuracy, we apply Chamfer Distance-L1 (C-L1), Accuracy (Acc.), Completion (Comp.), and F-score.

\textbf{Baselines}. Our main comparison is Panmap \cite{schmid2022panoptic}, against which we evaluate label accuracy in terms of panoptic, semantic, and instance, as well as geometry reconstruction quality. Panmap is a dense mapping system utilizing multiple TSDF maps with panoptic labels. However, it fails to address overlaps between submaps, and the panoptic label is determined by merging segmentation results without refinement. Additionally, we compare semantic label accuracy and geometry quality with Kimera~\cite{Rosinol20icra-Kimera}, a semantic mapping system that employs a single TSDF map.

\textbf{Implementation Details}. We use dataset-provided ground truth poses to avoid errors from biased pose estimates, ensuring direct comparisons. The input image panoptic segmentation is generated using Detectron2~\cite{wu2019detectron2} with its pre-trained weights. Depth information comes from depth images for the flat dataset and LiDAR point clouds for SemanticKITTI. We utilize a voxel size of 0.04m for flat dataset and 0.1m for SemanticKITTI dataset. The truncated distance for flat dataset is 0.08m and for SemanticKITTI dataset is 0.6m.
\begin{table}[t]
\centering
\tablestyle{8pt}{1.1}
\begin{tabular}{x{55}y{35}x{35}x{40}}
     \toprule
\textbf{Dataset}& \textbf{Method} & \textbf{mIoU} $\uparrow$ & \textbf{Accuracy} $\uparrow$ \\

     \midrule
 & Kimera & 0.338 & 0.770 \\
  \multirow{-2}{*}{\begin{tabular}[c]{@{}c@{}}Semantic\\KITTI\end{tabular}}  & 
  \cellcolor[HTML]{EFEFEF} DHP &\cellcolor[HTML]{EFEFEF}  0.263 & \cellcolor[HTML]{EFEFEF} 0.713 \\
       \midrule
 & Kimera & 0.670 & 0.852  \\
  \multirow{-2}{*}{flat}  & 
  \cellcolor[HTML]{EFEFEF} DHP & \cellcolor[HTML]{EFEFEF} 0.629 & \cellcolor[HTML]{EFEFEF} 0.854 \\
 \bottomrule
\end{tabular}
\caption{Comparison of semantic label accuracy between Kimera and DHP-mapping across two datasets.
}
\label{tab::kimera}
\vspace{-1.5em}
\end{table}
\begin{figure*}[h]
    \centering
    \includegraphics[width=1\linewidth]{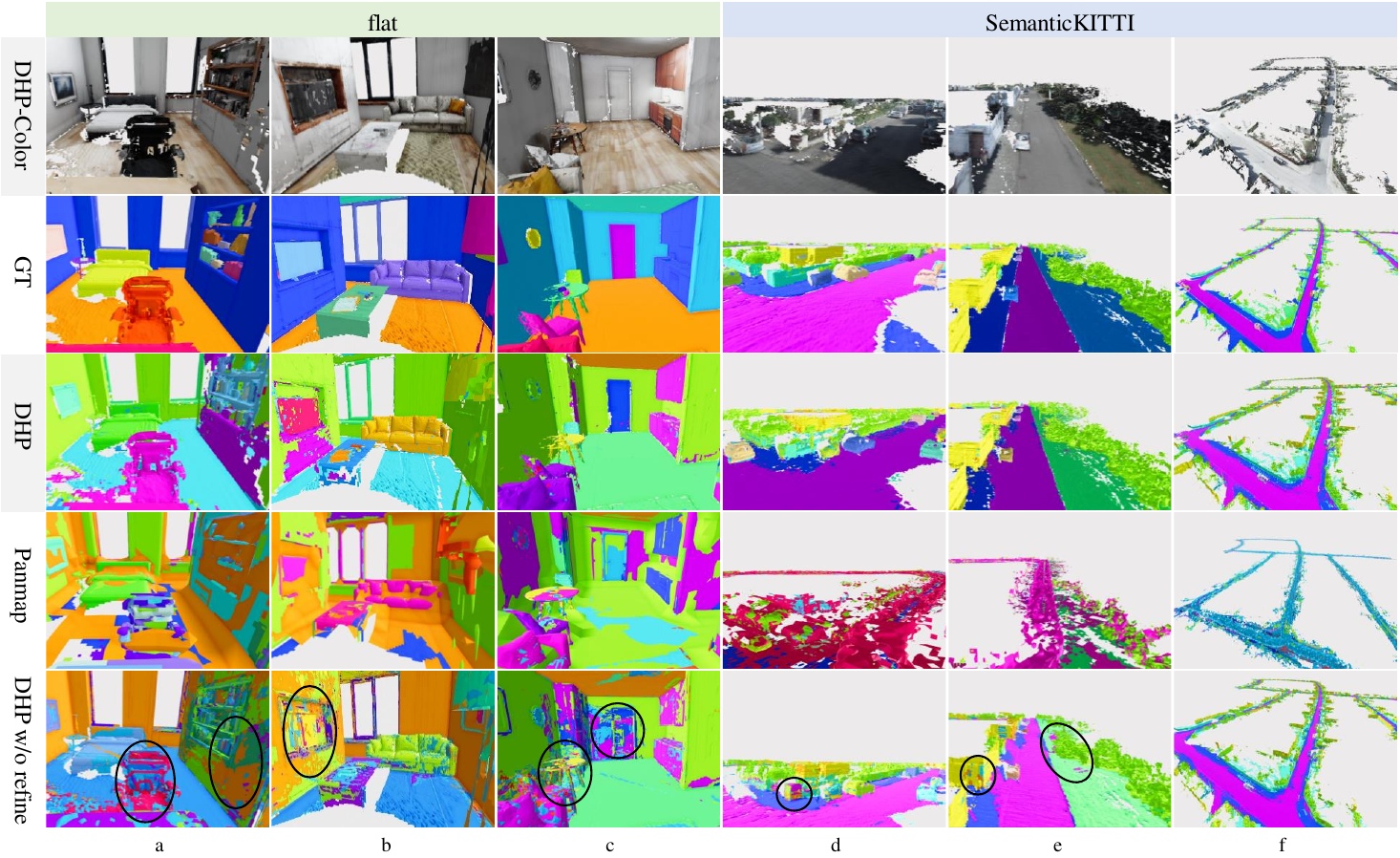}
    \caption{Visualization results of dense panoptic mapping systems run on flat and semanticKITTI datasets. Meshes are 
 extracted from the TSDF map using the marching cubes algorithm. The first line displays our system's map reconstruction results using the color values stored in their TSDF layers. The second to the fifth lines show the maps with label results. Different colors in each sub-figure represent unique object IDs. Compared with Panmap, our DHP-Mapping produces more consistent labels and reconstructs denser and more accurate geometry~(obvious in columns d-e-f). The proposed refinement techniques help to reduce submap overlaps and enhance label accuracy, without which submaps tend to intertwine with each other~(highlighted by black circles). }
    \label{fig:main_results}
        \vspace{-1em}
\end{figure*}
\subsection{Mapping Results Evaluation}
Our system focuses on building a map with dense geometry and rich semantics of environments, thereby boosting robots' situational awareness and making them perform more effectively.
In this section, we evaluate our mapping results utilizing established metrics. Our evaluation specifically seeks to address critical questions concerning labeling accuracy, geometric precision, impacts of our proposed refinement module, and advantages of our proposed map data structure. We recommend visiting our webpage ~\url{http://hutslib.github.io/DHP-Mapping} for more details.

\textbf{Does our system meet the requirement to reconstruct the scene with comprehensive and consistent labels?} 
We compare our system against Panmap~\cite{schmid2022panoptic} to assess label accuracy using panoptic, semantic, and instance metrics. The quantitative results, outlined in Table \ref{tab:main-eval}, reveal that our system significantly outperforms Panmap on flat dataset across all metrics and also exhibits an enhancement on SemanticKITTI dataset. 
The superior quantitative results demonstrate our system's advancement in comprehensively reconstructing scenes, accurately
categorizing semantic classes and tracking object IDs across frames. When compared with Kimera~\cite{Rosinol20icra-Kimera}, our system showcases competitive semantic labeling accuracy as results presented in Table~\ref{tab::kimera}. It's worth mentioning that Kimera focuses solely on semantic mapping and eschews the tracking of instance IDs. However, our system uses a more comprehensive label choice, incorporating both semantic categories and instance IDs. Our proposed system allows for a more in-depth environment representation, essential for robotic applications that require distinguishing different instances.

\begin{table}[t]
\centering
\tablestyle{6pt}{1.1}
\begin{tabular}{x{30}y{30}x{25}x{27}x{25}x{34}}
     \toprule
\textbf{Dataset}&\textbf{Method}& \textbf{Acc.}$\downarrow$ & \textbf{Comp.}$\downarrow$&\textbf{C-L1}$\downarrow$&\textbf{F-Score}$\uparrow$\\
     \midrule
 & Kimera & \textbf{0.0430} & {\ul 0.1871} & { \ul 0.1151} & {\ul 55.54} \\
  & Panmap & 0.0688 & 0.3119& 0.1904 & 9.16 \\
  \multirow{-3}{*}{\begin{tabular}[c]{@{}c@{}}Semantic\\KITTI\end{tabular}}  & 
  \cellcolor[HTML]{EFEFEF} DHP &\cellcolor[HTML]{EFEFEF} {\ul0.0506} & \cellcolor[HTML]{EFEFEF} \textbf{0.0584} & \cellcolor[HTML]{EFEFEF}  \textbf{0.0549} & \cellcolor[HTML]{EFEFEF}  \textbf{56.34} \\
       \midrule
 & Kimera & {\ul0.0076} & \textbf{0.0648} & \textbf{0.0362} & \textbf{89.99} \\
   & Panmap & 0.0086 & 0.0763& 0.0424 & 88.11 \\
  \multirow{-3}{*}{flat}  & 
  \cellcolor[HTML]{EFEFEF} DHP & \cellcolor[HTML]{EFEFEF} \textbf{0.0073} & \cellcolor[HTML]{EFEFEF}  {\ul0.0658} & \cellcolor[HTML]{EFEFEF} {\ul0.0365} & \cellcolor[HTML]{EFEFEF} { \ul 89.75} \\
 \bottomrule
\end{tabular}
\caption{Comparisons of geometry reconstruction quality between DHP-Mapping and SOTA metric-semantic mapping algorithms. We \textbf{bold} the best results and {\ul underline} the second best results. Note: Acc., Comp. and C-L1 are reported in meters.}
\label{tab:geometry}
\vspace{-2em}
\end{table}

\textbf{Depart from semantic reconstruction, how does our algorithm's performance on metric mapping?}
We access map geometry quality and benchmark our system's performance against SOTA metric-semantic mapping algorithms~\cite{schmid2022panoptic, Rosinol20icra-Kimera} using standard dense reconstruction metrics. The qualitative outcomes are detailed in Table \ref{tab:geometry}. It shows that our method produces a denser and more accurate map than Panmap on SemanticKITTI, highlighted by comparing the d, e, and f columns in Fig.~\ref{fig:main_results}. 
We attribute this to the different integration methods. Our system employs a ray-tracing method, updating all voxels from the sensor center to a truncation distance behind an observed point, whereas Panmap uses a projective method, only creating blocks at observed sparse LiDAR points' locations. This shows our system has superior generalization ability across both indoor and outdoor environments and is suitable to both RGB-D and LiDAR-camera settings. Moreover, compared to Kimera's single TSDF approach, our submap-based design is achieved without sacrificing much geometric accuracy. Instead, it manages the map in a hierarchical way and provides greater flexibility in accessing objects' geometric information.

\textbf{Does our label refinement algorithms enhance labeling quality?}
To show the effectiveness of the refinement module in our proposed system, which consists of the inter-submaps label management module and the CRF label optimization module, we conduct an ablation study by removing the refinement module from our mapping system. The quantitative results, denoted as ``DHP w/o refine'' in Table \ref{tab:main-eval}, indicate that the exclusion of the refinement module results in an obvious decline in performance. The visual comparisons in Fig.~\ref{fig:main_results} indicate that directly integrating imperfect panoptic segmentation results into the map leads to unclear submap boundaries, causing submaps to mix with others. The accuracy improvement is especially notable in indoor environments, where objects are closely positioned. Thus, in such settings, inaccurate segmentation will likely result in submap overlaps, making label accuracy benefit more from the refinement module.

\begin{figure}[t]
	\centering
	\subfigure[Original scenario]
	{\label{fig:before_manipulation}\centering\includegraphics[width=0.45\linewidth]{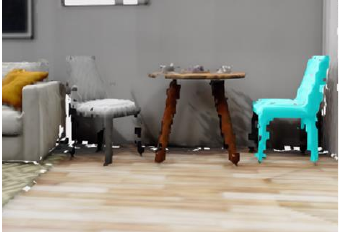}}
	\subfigure[Object manipulations]
	{\label{fig:after_manipulation}\centering\includegraphics[width=0.45\linewidth]{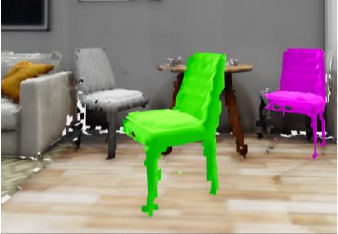}}
	\caption{An example of object manipulations using the submap data structure. We duplicate the blue chair-submap in the original scenario~(new submap shown in green) and change it to a new position~(new submap shown in pink). }	
	\label{fig:manipulation}
 \vspace{-0.5em}
\end{figure}  
\begin{table}[t]
\centering
\tablestyle{3pt}{1.1}
\begin{tabular}{x{75}x{70}x{70}}
     \toprule                                                               
\textbf{Scene~/~Voxel}& \textbf{Object~/~Voxel} & \textbf{Retrieval time}\\
     \midrule
      flat-01~/~4,099,223 & bed-a~/~61,745 & 21 milliseconds \\
  \begin{tabular}[c]{@{}c@{}}Semantic\\ KITTI-08\end{tabular}~/~82,529,555 & road~/~22,842,416 & 628 milliseconds\\
 \bottomrule
\end{tabular}
\caption{Objects retrieval time with single TSDF map structure. We show the voxel size of the entire map and the desired object. The data retrieval process is implemented in a single thread way and the program runs in a 12th generation i7 CPU with 2.3GHz.}
\label{tab:retrival}
\vspace{-2em}
\end{table}

\textbf{What advantages does hierarchical data structure with panoptic labels offer?}
As previous analysis highlighted, the use of a comprehensive labeling system in our mapping system greatly enriches scene representation. Besides, the hierarchical submap-based data structure facilitates submap-level manipulation and ensures rapid information retrieval. 
We show an example in Fig.~\ref{fig:manipulation}. In this demo, we add a new object to the scenario by directly duplicating an existing submap and adjusting the position of an object by modifying the relative pose between the submap and the global coordinate. This operation is feasible through our framework while unachievable with a single TSDF map structure.
Additionally, our system streamlines data access by allowing direct retrieval of all voxels associated with a specific object via submap querying. This eliminates the cumbersome need to search through every voxel to identify those matching the desired label. Table ~\ref{tab:retrival} gives an example of object retrieval time when using a flat single TSDF structure, while leveraging our hierarchical submap-based data structure the retrieval time can be ignored.
\section{Conclusion}
\label{sec:conclusion}
In this work, we design a dense volumetric mapping system that uses multiple TSDF submaps and panoptic labels to represent the scene hierarchically and holistically, while maintaining voxel-level and submap-level metric and label information.
The proposed inter-submaps label management module ensures the disjoint of spatial information in each submap. The CRF label optimization module improves the accuracy of panoptic labels by taking advantage of the inherent cohesion of objects and incorporating contextual information from the entire scene.
This hierarchical TSDF submaps with panoptic labels data structure enable high-level interactive tasks and dynamic environment modeling. 
In future work, more abstract and high-level representation can be further extracted and integrated above this data structure. 
This includes establishing topological connections between entities and integrating language-based expressions that go beyond metric and symbolic representations.

\bibliographystyle{IEEEtran}
\bibliography{IEEEabrv,ref}

\end{document}